\documentclass[conference]{IEEEtran}
\ifCLASSINFOpdf
\usepackage[pdftex]{graphicx}
\else
\fi
\usepackage{amsmath}
\usepackage{mathtools}
\usepackage{amssymb}

\DeclareMathOperator*{\argmin}{arg\,min}

\newcommand{\Tff}{\ensuremath{\Theta_\mathrm{ff}}}
\newcommand{\Tbp}{\ensuremath{\Theta_\mathrm{bp}}}  

\begin{document}
%
\title{A Spiking Network for Inference of Relations Trained with Neuromorphic Backpropagation}

\author{\IEEEauthorblockN{Johannes C. Thiele, Olivier Bichler, Antoine Dupret}
\IEEEauthorblockA{CEA, LIST \\
Gif-sur-Yvette CEDEX, France\\
johannes.thiele@cea.fr}
\and
\IEEEauthorblockN{Sergio Solinas, Giacomo Indiveri}
\IEEEauthorblockA{Institute of Neuroinformatics, \\ 
University of Zurich and ETH Zurich \\
Zurich, Switzerland}


}


%


\IEEEoverridecommandlockouts
\IEEEpubid{\makebox[\columnwidth]{Accepted as a conference paper at IJCNN 2019 \copyright 2019 IEEE.  \hfill} \hspace{\columnsep}\makebox[\columnwidth]{ }}

\maketitle

\begin{abstract}
The increasing need for intelligent sensors in a wide range of everyday objects requires the existence of low power information processing systems which can operate autonomously in their environment. In particular, merging and processing the outputs of different sensors efficiently is a necessary requirement for mobile agents with cognitive abilities. In this work, we present a multi-layer spiking neural network for inference of relations between stimuli patterns in dedicated neuromorphic systems. The system is trained with a new version of the backpropagation algorithm adapted to on-chip learning in neuromorphic hardware: Error gradients are encoded as spike signals which are propagated through symmetric synapses, using the same integrate-and-fire hardware infrastructure as used during forward propagation. We demonstrate the strength of the approach on an arithmetic relation inference task and on visual XOR on the MNIST dataset. Compared to previous, biologically-inspired implementations of networks for learning and inference of relations, our approach is able to achieve better performance with less neurons. Our architecture is the first spiking neural network architecture with on-chip learning capabilities, which is able to perform relational inference on complex visual stimuli. These features make our system interesting for sensor fusion applications and embedded learning in autonomous neuromorphic agents.
\end{abstract}


%
\IEEEpeerreviewmaketitle

\section{Introduction}

Artificial neural networks are mostly known for their feedforward processing capacities, where a well-defined input $X$ leads to a well-defined output $Y$. This is however different to the processing paradigm of the human brain. Even pathways which are often characterized by their feedforward structure, such as the visual system, possess a large level of recurrent connectivity between different levels of processing \cite{Binzegger:2004}. This type of recurrent connectivity allows the brain to infer not only in a one-directional feedforward fashion, but in a associative, relational way, potentially involving several hierarchical levels \cite{Felleman:1991}. Stimuli which are related to each other produce correlated activity patterns, and presentation of any of these stimuli will also produce high activity in an area representing other, related stimuli. This enables neural networks to perform relational inference, i.e.\ to relate different stimuli with each other in a multi-directional way.

The field of neuromorphic engineering aims at building dedicated brain-inspired computing architectures to harness beneficial computational properties of the brain \cite{Chicca:2014}. In this work, we present an approach to train a neuromorphic network of spiking neurons with the aforementioned multidirectional inference capabilities. In previous work on biologically inspired implementations of relational networks \cite{Deneve:2001}\cite{Diehl:2016}\cite{Thiele:2017WSA}, connections are either hardwired or were learned with biologically inspired learning rules, such as spike-timing-dependent plasticity (STDP) \cite{Bi:1998}. In this paper we tackle this problem from a more practical point of view. In particular, we drop some biological constraints to improve the training of the network, while keeping those aspects that are beneficial for an energy efficient neuromorphic implementations. 

For this purpose, we use a special version of the backpropagation algorithm outlined in \cite{Thiele:2018c}, which is adapted to the event-based communication constraints found in neuromorphic systems. The communication of gradient signals is performed via the propagation of spike events. Although our framework compromises some aspects of biological plausibility, such as requiring weight symmetry, it is fully compatible with most aspects of the spike-based neuromorphic computing paradigm. Communication of forward and backward propagated information is fully based on signed binary events, without the need to process floating point numbers, in contrast to the standard backpropagation algorithm \cite{Rumelhart:1986b}. This makes it particularly suitable for integration in digital neuromorphic platforms (such as \cite{Frenkel:2018}).

Our work can be seen as an extension to previous approaches for training relational networks of spiking neurons. We demonstrate that our algorithm achieves superior performance compared to the biologically inspired approaches \cite{Deneve:2001}\cite{Diehl:2016}\cite{Thiele:2017WSA}. Additionally, we show that our network is able to deal with more complex visual stimuli and set them in relationship with each other. This is demonstrated by learning relational inference on the visual XOR task, using images from the MNIST dataset \cite{LeCun:1998}. 

The potentially low energy processing mechanism of training and inference could make our systems potentially attractive for sensor fusion applications on mobile and embedded platforms. 

\section{Methods}

In this work, we use the simple integrate-and-fire (IF) neuron model with threshold $\Tff$ which is commonly used in theoretical work on spiking neural networks, in particular digital implementations:
\begin{equation} \label{update_potential}
V_i^l(t) =  V_i^l(t-\Delta t) - \Tff s_i^l(t-\Delta t) + \sum_j w^{l}_{ij}s^{l-1}_j(t) + b^l_i(t).
\end{equation}
The dynamics in this section are described in the framework of a time-stepped simulation with time step $\Delta t$. The variable $b^l_i(t)$ represents an optional bias value, which is only added once to the integration variable $V_i^l(t)$ of the neuron when a new stimulus is presented and is otherwise zero. The arrival of spikes is described by a spike activation function  $s^l_i(t) \in \{-1,0,1\}$. This function is a threshold function which is mostly $0$, except when the integration variable crosses the negative or positive threshold. In this case, a signed spike postsynaptic event is triggered:
\begin{equation} \label{spike_activation_function}
s^l_i\left(V_i^l(t)\right) = \begin{cases}
	\; 1 &  \mathrm{if}\;V^l_i(t)>\Tff \\
	\; -1 &  \mathrm{if}\;V^l_i(t)<-\Tff\;\mathrm{and}\; x^l_i(t)>0\\
	\; 0 & \mathrm{otherwise} 
	\end{cases}.
\end{equation}
Note that in the description here, $s^l_i(t)$ will be referred to as a ternarized variable, although the propagated spike signal in an event based implementation is binary (since $s^l_i(t)=0$ simply corresponds to no spike being triggered). After a neuron has fired, its integration variable is decremented or incremented by the threshold value $\Tff$, according to the value of $s^l_i(t)$. This reset is  represented by the second term on the r.h.s.\ of \eqref{update_potential}. The reason why this neuron model is particularly attractive for digital implementations is the fact that integration and neuron dynamics are completely multiplication free, and only based on accumulations and comparisons of state variables.

In contrast to most implementations of spiking neural networks, we allow for the propagation of negative spikes. This is because our algorithm will try to approximate the accumulated response of the neuron as a standard, floating point based neuron from an artificial neural network. If we restrict the model to positive spikes, the output of the neuron has a strong dependence on the order of spike arrival. To give an example, if there are two spikes which arrive at the neuron, one of them increasing $V_i^l(t)$ by $1$ and the other one decreasing $V_i^l(t)$ by $-1$, the net change of the integration variable will be $0$. However, if the positive contribution is integrated first and the integration passes the threshold value, the neuron will emit a spike and be reset, which would not be the case if the negative contribution is integrated first. By allowing negative spikes, spikes propagated in excess can be corrected by a subsequent negative spike if the integration drops below the negative threshold $-\Tff$ (and the total integration becomes smaller than 0). This way the total propagated output of a neuron will be approximately proportional to the total integrated input, and therefore represents approximately a linear activation function.

Note that the spike activation function \eqref{spike_activation_function} is asymmetric with respect to positive and negative spikes. For negative spikes, it is conditioned on a trace $x^l_i(t)$ which accumulates spike information over longer time windows: 
\begin{equation} \label{learning_rate_trace}
x_i^{l}(t) = \eta s_i^{l}(t) + x_i^{l}(t-\Delta t).
\end{equation}
The trace is updated every time a postsynaptic spike is triggered in a neuron. Since $s^{l}_i(t)$ is a ternary variable, the trace is simply an weighted activity counter. In the spike activation function \eqref{spike_activation_function}, it ensures that the total output propagated by a neuron is always larger or equal to 0. The neuron model can this way be seen as an approximate implementation of the Rectified Linear Unit (ReLU) activation function. If $x_i^{l}(t) \leq 0$, that means if the total propagated output is negative, no negative spike can be triggered. This ensures that the total propagated output is always positive. This trace will also play a role in the weight update rule which is derived later, where it will serve as a representation of the total received input by a synapse. This explains why the trace is already weighted by the learning rate $\eta$. For the purpose of restricting the output to positive values, this factor is irrelevant since it only scales the variables without changing its sign.

\begin{figure*}
	\begin{centering}
		\includegraphics[width=1\textwidth]{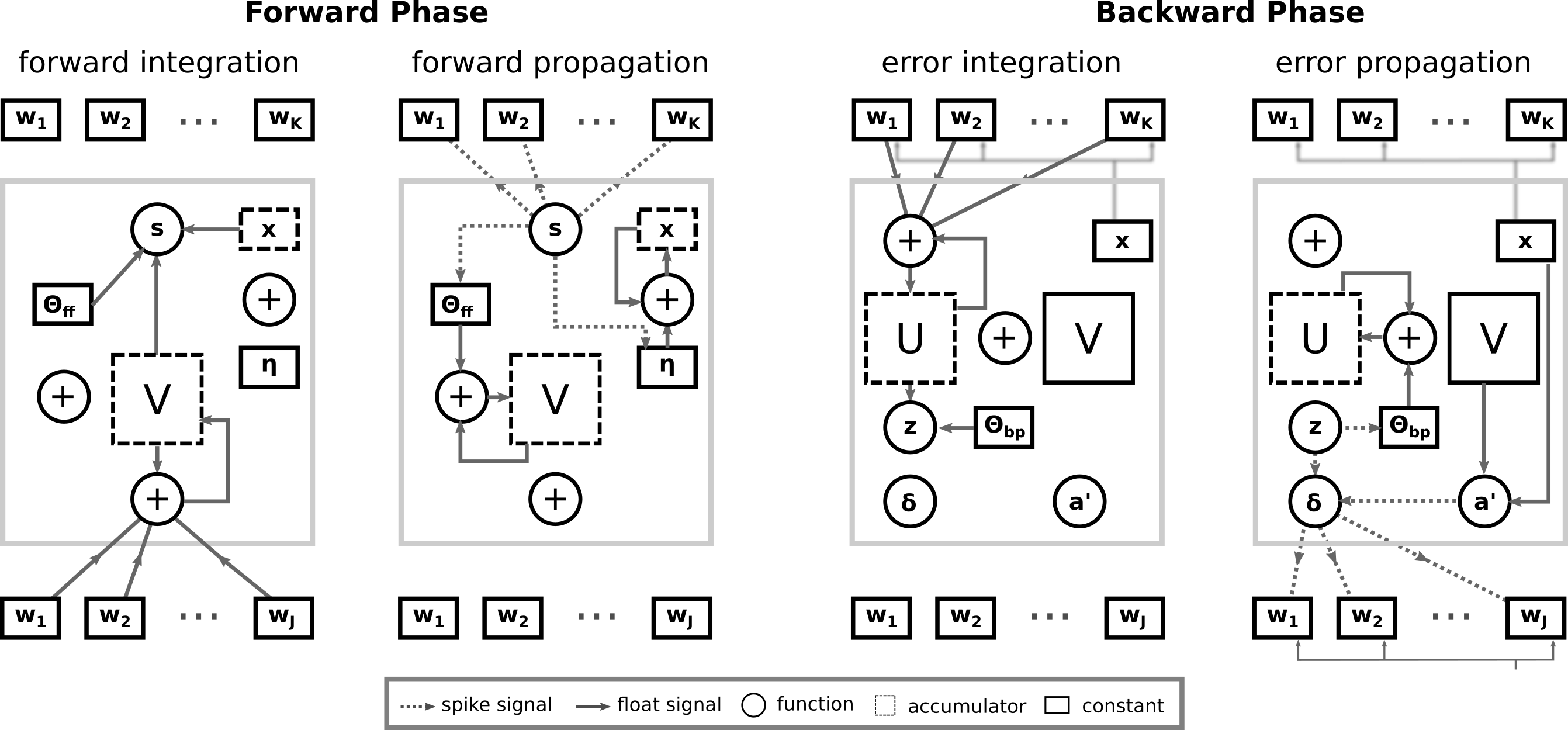}
		\par\end{centering}
              \caption{Forward and backward phase in a single hidden neuron. Ternary signals (i.e.\ binary spikes) are represented by dashed lines and are multiplied with their targets. 
              \textbf{Forward integration:} Every time a spike signal arrives at one of the synapses $\{w_1,...,w_J\}$, their positive or negative value is added to the integration variable  $V$ \eqref{update_potential}, depending on the sign of the spike. After each such event the integration variable is compared to the threshold value $\pm \Tff$ and the synaptic trace $x$ by the function $s$ \eqref{spike_activation_function}, which decides if a spike is triggered or not. \textbf{Forward propagation:} If the conditions imposed by $s$ are satisfied, a postsynaptic spike is triggered. This spike increases the trace $x$ \eqref{learning_rate_trace} by the learning rate $\pm \eta$. Additionally, it is send to the outgoing connections. The signal also applies $\pm \Tff$ to $V$ depending on the sign of $s$ \eqref{update_potential}.
              \textbf{Error integration:} Signed error spikes are received through the weights $\{w_1,...,w_K\}$ of the outgoing connections. Depending on their sign, they add their positive or negative value to the error integration variable $U$ \eqref{update_error}.  After each such signal, $U$ is compared to $\pm \Tbp$ by the function $z$ \eqref{error_ternarization}. \textbf{Error propagation:} If the threshold is crossed, a signed error spike is emitted. $U$ is incremented by $\Tbp$ or $-\Tbp$. This ternarized signal is gated by the surrogate activation function derivative \eqref{straight_through_estimator_x}, which is calculate based on $V$ and $x$, and propagated through the incoming connections. The weights of the neuron are updated \eqref{weight_update_final} using this error signal and the trace \eqref{learning_rate_trace} from the neuron in the layer below.}
	\label{BP_process}
\end{figure*}

\subsection{Spiking backpropagation by dynamic error ternarization}

The first obstacle in the derivation of a backpropagation algorithm for spiking networks is the non-differentiability of the activation function with respect to the input. This is because  $s_i^{l}(t)$ is a temporally discontinuous variable. We therefore cannot formally take the derivative of the activation function with respect to the neuron variables, but instead have to use a surrogate derivative which describes in a meaningful way how the activation of a neuron changes as a function of its input and weights.

Several solutions have been proposed in this context (see for instance \cite{Lee:2016}\cite{Yin:2017}\cite{Neftci:2017}\cite{Wu:2018SpikingBP}\cite{Huh:2017}\cite{Tavanaei:2018}\cite{Zenke:2018SuperSpike}. We use in this work a similar approach to \cite{Yin:2017} using an approximate surrogate activation function derivative which is calculated depending on the accumulated activity of the neuron:
\begin{equation} \label{straight_through_estimator_x}
a_i^{'l}(t) \coloneqq \begin{cases}
	\; 1 &  \mathrm{if}\; V^l_i(t) > 0 \;\mathrm{or}\;x_i^l(t)>0 \\
	\; 0 & \mathrm{otherwise}
	\end{cases}.
\end{equation}
This corresponds to the derivative of a ReLU activation function of the total received input. The condition on $V^l_i(t)$ is included only for the case where $x_i^l(t) = 0$, to be able to propagate gradients even if no spike is triggered yet.

The next step is to convert the accumulated errors in each neuron to signed events. For this purpose, we introduce a second compartment with a threshold $\Tbp$ in the neuron which integrates error signals analogously to \eqref{update_potential}:
\begin{equation} \label{update_error}
U_i^l(t) =  U_i^l(t-\Delta t) - \Tbp z_i^l(t-\Delta t) + \sum_k w^{l+1}_{ki}(t)\delta^{l+1}_k(t).
\end{equation}
This equation is used to trigger a ternary variable which is defined in analogy to \eqref{spike_activation_function}:
\begin{equation}\label{error_ternarization}
z_i^{l}(t) = \begin{cases}
	\; 1 &  \mathrm{if}\; U_i^{l}(t) > \Tbp \\
	\; -1 &  \mathrm{if}\; U_i^{l}(t) < -\Tbp \\
	\; 0 & \mathrm{otherwise}
	\end{cases}.
\end{equation}
This discretizes in a dynamics fashion the current integrated error.
The basic process is the same as for the discretization of the input signal into a sequence of binary signals for forward processing. Forward and backward propagation are therefore highly symmetric in their corresponding mechanisms.

%

The most important difference between forward and backward pass is that the error signal has to be multiplied by the derivative of the activation function in each neuron. The error for further propagation is then obtained by gating the ternarized variable $z_i^{l}(t)$ with the surrogate activation derivative derivative:
\begin{equation}\label{ternarized error}
\delta_i^{l}(t) = z_i^{l}(t) a^{'l}_i(t),
\end{equation}
This error signal will remain a ternarized variable if $a_i^{'l}(t)\in\{-1,0,1\}$, which is the case for \eqref{straight_through_estimator_x}. This signal is backpropagated through the synapses like a spike event to calculate the error in the other layers of the network in the same fashion. Additionally, it is applied in the weight update rule: 
\begin{equation}\label{weight_update_final}
\Delta w_{ij}^{l}(t) = \begin{cases}
	\; -x_j^{l-1}(t) &  \mathrm{if}\; \delta_i^{l}(t) = 1 \\
	\; x_j^{l-1}(t) &  \mathrm{if}\; \delta_i^{l}(t) = -1 \\
	\; 0 & \mathrm{otherwise}
	\end{cases},
\end{equation}
In a fully asynchronous implementation, this update is performed every time an error spike signal passes a synapse. For an implementation which is closer to standard backpropagation, all weight updates are accumulated before they are applied to each weight after the backpropagation phase has finished.

Implementing the backpropagation of errors and the weight update in the way described in this section leads to a learning algorithm which, exactly as the forward propagation, only involves additions and comparisons of floating point numbers. The error signal is generated in the output layer and propagated continuously over a certain time span, just like during forward propagation. This enables us to obtain an error signal with  dynamic precision. By encoding the error into more spikes and performing error propagation for a longer time, the precision of the propagated error can be increased. Figure \ref{BP_process} provides a visual description of the algorithm. In our simulations we found it useful to implement learning rate decay, which can easily be included in our formalism by changing the value of the increment $\eta$.

Several implementation possibilities exist depending on the desired substrate and hardware constraints. A major advantage of our implementation is that it is fully based on accumulations and no explicit spike time has to be communicated for computation (only if a spike arrives at a certain moment in time or not). This removes the need to propagate and save timestamps. The spike signal can directly be propagated as a binary signal which encodes the sign of the event. A digital implementation can additionally profit from the fact that all required operations are either sums or comparators. In the case of an analog implementation, the neuron and synapse models probably have to be adapted to reflect more accurately the dynamics of analog neurons. It can however also profit from the low precision requirements of spike propagation, since the signed spike event could be encoded as a bipolar signal of arbitrary amplitude.

\subsection{Loss function}

For our experiments, each branch of the relational network will be trained to predict a desired spike output pattern of a variable, given as an input the other two. We use the simple squared $L_2$ loss function:
\begin{equation}\label{loss_function}
\mathcal{L} =  \sum_i \frac{1}{2}(y_i-t_i)^2
\end{equation}
where $y$ is the accumulated activity of the inferring population and $t_i$ the target spike pattern activity. We therefore use a coding were information is encoded by firing rates. In our implementation, we use $y_i = x_i^L+V_i^L$ to enable learning even if there is no spike in the final layer. This loss gives as an error for the final layer at the beginning of the backpropagation phase:
\begin{equation}
\frac{d\mathcal{L}}{dy_i} = (y_i-t_i).
\end{equation}
This error can thus be calculated without multiplications. In the inferring layer, it is directly transferred to the error integration variable $U_i^L$ of each neuron in the final layer. Note that this way it is possible that, $|U_i^L| > \Tbp$. In our implementation, the neuron will produce a spike at every time step and decrease $U_i^L$ by $\Tbp$ according to \eqref{update_error} as long as the integration variable exceeds the threshold. This allows us to propagate a high precision gradient which is discretized into binary spike events. In principle it is also possible to use another spike conversion scheme, as it is often done for the conversion of the real valued input to spike trains. The advantage of our conversion scheme is that it does not change the scale of the error, besides the approximation introduced by the spike quantization, which effectively restricts the error to integer values.

\subsection{Weight initialization}

Due to the similarity of the IF neuron with the ReLU activation function, we use a initialization method proposed for deep networks using this type of activation function \cite{He:2015}.
\begin{equation}
\mathbf{w} \sim \mathcal{N}(0, \sqrt{2/n^\mathrm{in}}),
\end{equation}
where $n^\mathrm{in}$ is the number of incoming connections of the neuron.

\section{Relational network}

We now describe the relational network topology, which will be the network trained with the neuromorphic backpropagation algorithm. The relational network allows to fuse several inputs and set them in relation with each other. Based on a subset of input stimuli, the network is able to recreate the missing ones in the form of artificial patterns which possess a similar statistic as the original ones. The basic relational network used in this work can in principle be extended to arbitrarily complex structures by coupling several of these basic networks, since any of the populations can simply be replaced by another relational network.  


%
\begin{figure}
	\begin{centering}
		\includegraphics[width=1\columnwidth]{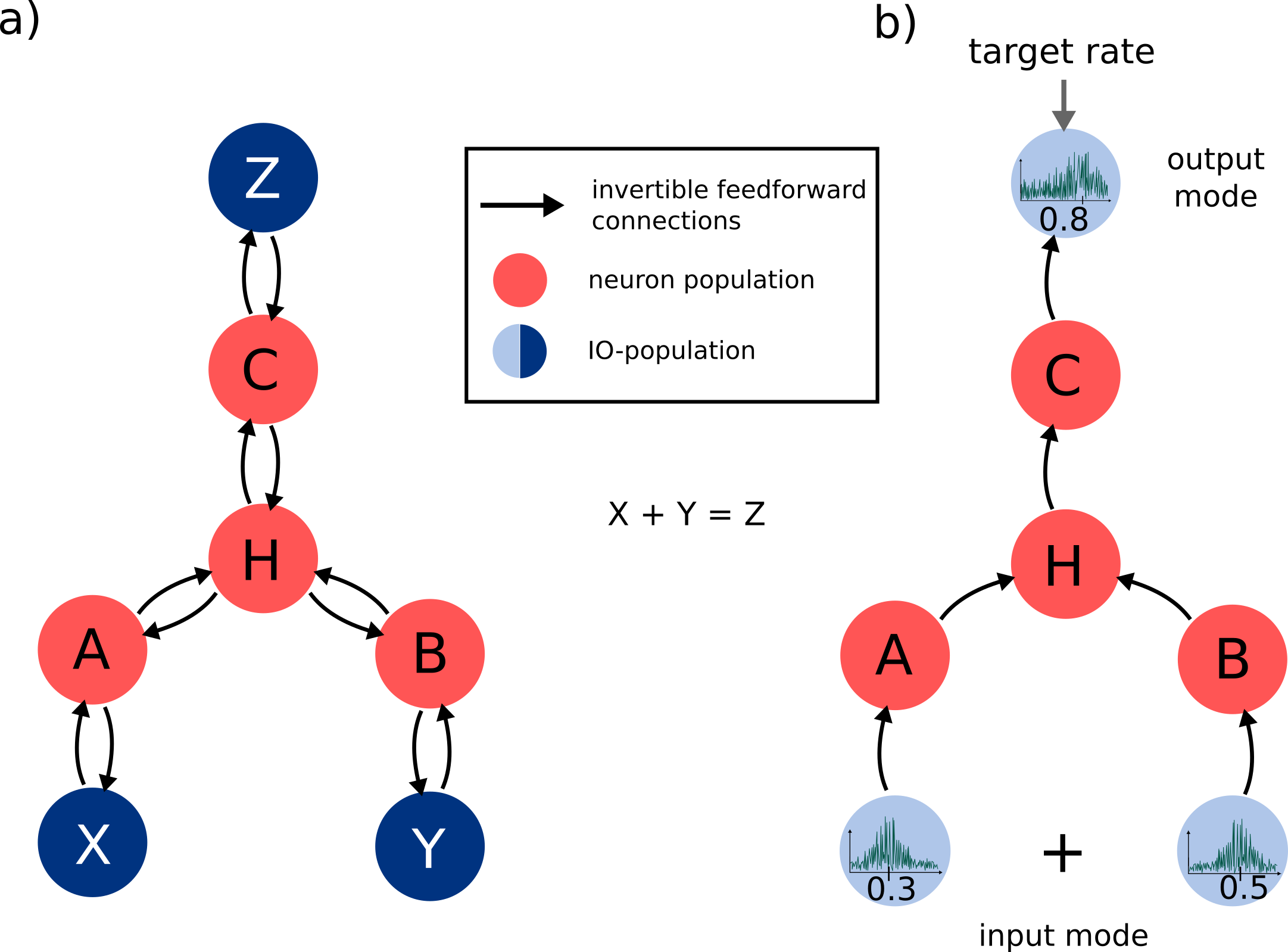}
		\par\end{centering}
              \caption{\textbf{a)} Relational network architecture with three variables. IO-populations are labeled \textbf{X}, \textbf{Y}, \textbf{Z}. Peripheral populations are labeled \textbf{A}, \textbf{B}, \textbf{C} and the hidden population \textbf{H}. \textbf{b)}: Training of the network. During learning, two populations serve as input populations and provide a firing pattern. The third population is trained to reproduce a target firing pattern based on this input. The roles of the  populations are interchanged during training to enable all variables to perform inference.}
	\label{network_learning}
\end{figure}

To represent a relation with $n$ variables in a relational network, we couple $n+1$ populations. $n$ of these populations will learn to represent the $n$ input variables, while population $n+1$ will set these represented variables in relationship with each other. We distinguish 3 types of populations (see figure \ref{network_learning}):
\begin{itemize}
\item input-output (IO) populations: These populations provide the input patterns to the network or reconstruct the missing input, depending on the inference direction
\item peripheral populations: These populations process the input from the IO-populations and hidden populations to find a high level representation
\item hidden populations: The hidden populations receive the processed input from the peripheral populations, processes it further and sends it to other peripheral populations. It therefore sets the different inputs from the peripheral populations in relation with each other.
\end{itemize}

Our network has the same inter-population connectivity structure as the relational network in \cite{Diehl:2016}, with the difference that additionally the IO-populations have feedforward input from the peripheral populations. The peripheral populations receive connections from their corresponding IO-populations and the hidden population. The hidden population receives feedforward input from all peripheral populations. In contrast to the network in \cite{Diehl:2016}, all connections are bidirectional in the sense that they allow the backpropagation of an error signal. The overall topology remains however equivalent regarding which populations with each other. Our architecture requires no recurrent connections between neurons in a population.

\subsection{Training procedure}

The network is trained in a supervised fashion, in the sense that a subset of the variables of the relation are provided as an input, while all other variables serve as targets. $n-1$ of the IO-populations provide input spike trains, while the remaining population is trained to reproduce the spike pattern of the missing variable (figure \ref{network_learning}). A subset of all possible connections is enabled such that for each variable the network functions as a feedforward network. This mechanism is rotated such that each input population serves as output populations equally often during training

While training the network on each processing direction of the relation, a certain subset of all existing synaptic connections have to be disabled. This allows the network for each direction to be in a pure feedforward mode.  At the same time, all relations are simultaneously presented in the network weights. All feedforward connections leading to the hidden population are simultaneously optimized for all inference directions of the relations (for instance in figure \ref{network_learning}, the weights XA, AH are optimized for inference of Y and Z, YB, BH for inference of X and Z and ZC, CH for inference of X and Y). 

For backpropagation, we let the error propagate through the network for a certain number of time steps. The inferring IO-population simply discretizes the error into subsequent spikes (one spike per time step) until their error integrator is below the threshold value.

\subsection{Input encoding}\label{Input encoding}

The rate is converted to a deterministic spike train with equally sized inter-spike intervals. The set of input spike times for a neuron in the interval $[t_\mathrm{start},t_\mathrm{stop}]$ during which an example is presented is then defined by:
\begin{equation}
\mathcal{T}_i = \{t_\mathrm{in}|\; t_\mathrm{in} = t_\mathrm{start} + a \frac{1}{r_i}  ,\; t_\mathrm{in} \leq t_\mathrm{stop},\;a\in \mathbb{N}\}.
\end{equation}
For the neurons in the IO-populations, the spike behavior in a time stepped simulation with time step $\Delta t$ is therefore imposed as:
\begin{equation}
s_i(t) = \begin{cases}
	\; 1 &  \mathrm{if}\; \exists\ t_\mathrm{in}\in \mathcal{T}_i :\;t-\Delta t < t_\mathrm{in} \leq t \\
	\; 0 & \mathrm{otherwise}
	\end{cases} 
\end{equation}
This deterministic coding is used to facilitate the representation of the numbers, but our experiments show that the scheme also works well with a more stochastic type of coding, as long as the firing rate is representative of the pixel value. A typical firing rate pattern induced in a neuron population by this kind of coding can be seen in figure \ref{firing_profiles}. 

Note that we used an explicit time in the definition of the spike input pattern. Since the network dynamics is defined without an explicit notion of a time, this time can simply be seen as a variable which describes the discretization density of the input spike pattern. For a static pattern, which is used for the demonstrations in this work, this discretization is seems kind of artificial. It is used here to reflect the fact that in the general case, the input will be arriving from a sensor which produces a dynamic number of spike events, which arrive at different points in time.

If the IO-population is used as an output population during inference, this mechanism is disabled and the population receives only input from the corresponding peripheral population via the incoming synapses.

\section{Experiments}

We now train the relational network topology on two relational inference tasks: A periodic addition and a visual XOR task. If not otherwise stated, we use the parameters given in table \ref{parameters}. Since our simulation does not require the definition of an explicit time scale, all time related variables are given in relative units.

All simulations were performed with an extended version of the N2D2 open source machine learning library \cite{Bichler:2017}.
\begin{table}
\renewcommand{\arraystretch}{1.3}
\caption{Parameters used for network training}\label{parameters}
\begin{center}
\begin{tabular}{c|c|c}
\hline
\textbf{Symbol} & \textbf{Description} & \textbf{Value} \\ 
\hline
$\Tff$ & Threshold for forward propagation & $1.0$ \\
$\Tbp$ & Threshold for backpropagation & $1.0$ \\
$\eta$ & Learning Rate & $0.00005$ \\
$r_\mathrm{max}$ & Maximal input firing rate & $0.12$ \\
$t_\mathrm{expl.}$ & Example presentation time & $100$ \\
$t_\mathrm{BP}$ & BP presentation time & $10$ \\
$\Delta t$ & Simulation time step & $1$ \\
$N_\mathrm{train}$ & \# relation samples used for training & $10000$ \\
$N^\mathrm{A,B,C}$ & \# neurons in peripheral populations & $256$ \\
$N^\mathrm{H}$ & \# neurons in hidden population & $128$ \\
\hline
\end{tabular}
\end{center}
\end{table}

\subsection{Implementing periodic addition}

\begin{figure*}
	\begin{centering}
		\includegraphics[width=\textwidth]{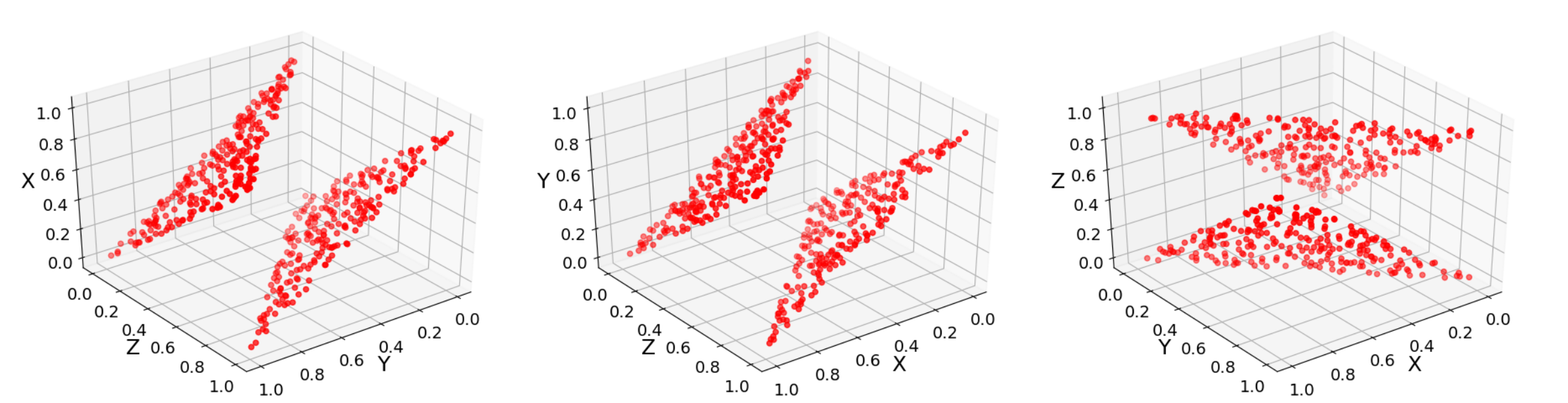}
	\par\end{centering}
              \caption{Relational inference for all three output variables. The two bottom axes describe the two input variables, while the vertical axis is the value inferred using \eqref{infer_variable}. We can observe that the network has learned to represent all possible directions of the relation accurately. The root-means-squared error (RMSE) averaged over the three populations is $0.0154$}
	\label{inference}
\end{figure*}
\begin{figure*}
	\begin{centering}
		\includegraphics[width=\textwidth]{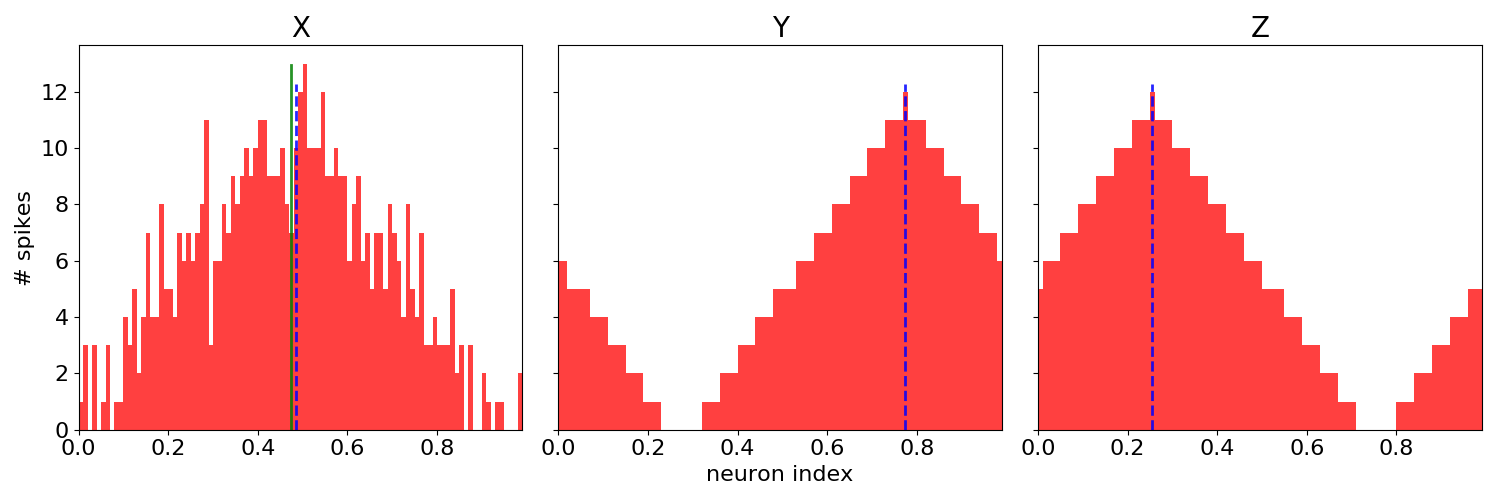}
		\par\end{centering}
              \caption{Firing rate profiles produced by the different populations during inference. Y and Z have firing rate profiles imposed to represent the two input variables, while X produces a firing rate profile based on this input, which presents the third number of the relation. The blue vertical lines represent the target numbers, the green line the number inferred by the population (based on \eqref{infer_variable}). The inferred profile of X is slightly noisier than the input firing rate profiles, but it presents accurately the desired profile (correct inferred value and overall shape)}
	\label{firing_profiles}
\end{figure*}

As a first demonstration of the architecture we implement an addition with periodic boundaries:
\begin{equation}\label{mod_addition}
\gamma = \alpha + \beta - \left \lfloor{\alpha + \beta}\right\rfloor.
\end{equation}
For implementing relations between numbers, we use a similar approach as \cite{Diehl:2016} and encode each number as a firing rate profile of the input population. The numbers $\alpha, \beta, \gamma \in [0,1]$ represent the numbers encoded by the IO-populations X, Y and Z. Each variable $\xi \in \{\alpha, \beta, \gamma\}$ is converted into a spike profile by assigning each of the $N$ neurons of an IO-populations a constant firing rate based on its index $i$: 
\begin{equation}
r_i(\xi) = r_\mathrm{max}\cdot|1-2|\xi-\frac{i}{N}|| 
\end{equation}
These rates are converted into spike trains using the method outlined in section \ref{Input encoding} (see figure \ref{firing_profiles} for a visualization). We use an IO-population size of $100$ neurons, which allows us to represent $\xi$ up to a theoretical precision of $\pm 0.005$.

Inference of the estimated value $\hat{\xi}$ is performed based on the firing pattern of the inferring population. For the visual inference task, the firing rate pattern of the inferring IO-population is directly taken (after a rescaling) as pixel values of the inferred image. The inferred number $\hat{\xi}\in [0,1]$ is derived from the firing rate pattern of the $N$ neurons by finding the neuron which minimizes the activity \eqref{learning_rate_trace} weighted distance to all other neurons in the same population:
\begin{equation}\label{infer_variable}
\hat{\xi} = \frac{1}{N}\argmin_i \sum_j x_j d_N[i,j])  
\end{equation}
using the periodic distance function:
\begin{equation}
d_N[i,j] \coloneqq \begin{cases}
	\; |i-j| &  \mathrm{if}\; |i-j| \leq N/2  \\
	\; N - |i-j| &  \mathrm{if}\; |i-j| > N/2
	\end{cases}.
\end{equation}

The response plots in figure \ref{inference} show that the network learns to accurately represent the relation. Each of the two input populations can be used to infer the value of the variable represented by the third population. With $0.0154$, the root-means-squared error (RMSE) is considerably lower than the errors obtained by \cite{Diehl:2016} and \cite{Thiele:2018}. Additionally, the network can reproduce the approximate firing rate profile of the encoded input, since each IO-populations was explicitly trained to do so (\ref{firing_profiles}). This is means the network can in particular reproduce an output which has the same scale as the original input.

\subsection{Visual XOR}

\begin{figure*}
	\begin{centering}
		\includegraphics[width=\textwidth]{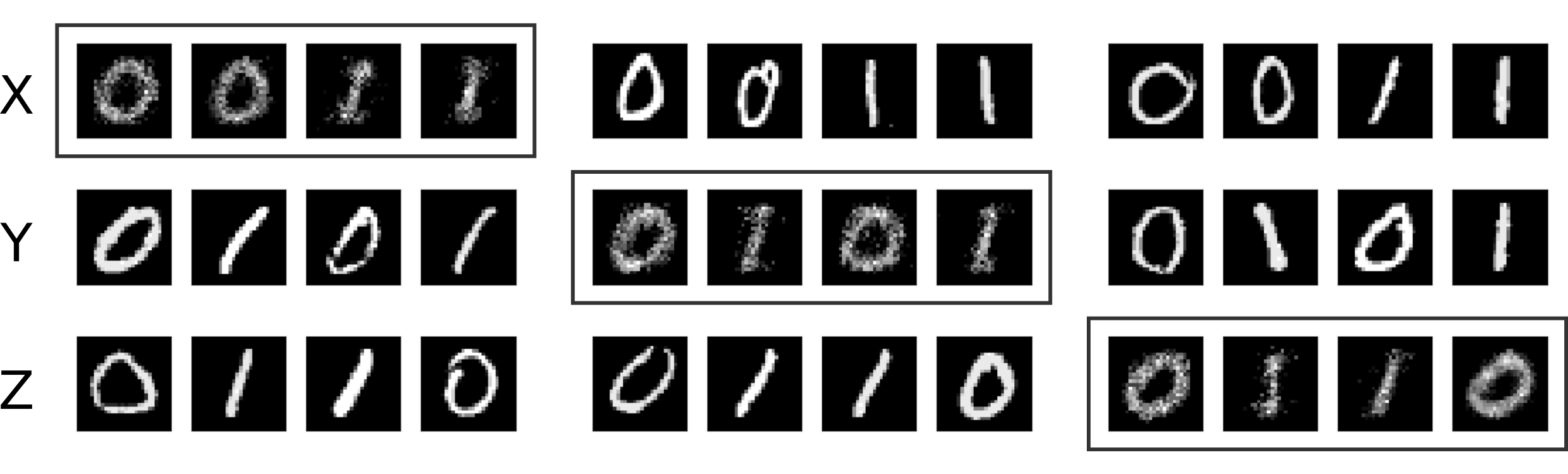}
		\par\end{centering}
              \caption{Visual relational inference for all three variables of relation \eqref{visual_xor}. The inferred stimuli are marked by boxes. We can observe that the network has learned to represent all possible directions of relation \eqref{visual_xor} accurately and it is able to produce artificial stimuli which are similar to the stimuli presented to the network during training.}
	\label{visual_inference}
\end{figure*}

We now apply the network to a more challenging task, which requires the network to  find abstract representations of the input data before setting them in relation with each other. For this purpose, we let the network learn the visual XOR task. The inputs X, Y, Z are now spike-encoded examples of the MNIST dataset representing the handwritten digits 0 and 1:
\begin{equation}\label{visual_xor}
l(\mathrm{Z}) = l(\mathrm{X}) \oplus l(\mathrm{Y}).
\end{equation}
The function $l \in \{0,1\}$ is the function which maps the images encoded by the IO-populations X, Y and Z to their corresponding labels. We chose in this case only examples of the dataset that represent the numbers $1$ and $0$, which will represent the boolean values \textrm{true} and \textrm{false} respectively. Note that this can be also seen as an addition $\mathrm{mod}\; 2$ of integers with 0 and 1 as possible values and is therefore in a sense similar to the previous task.

The samples used for training are randomly selected images from the MNIST training set of 60000 digits (numbers 0 to 9), with the condition that they are consistent with the relation. This means that the number of possible samples is extremely large (in the order of $4\cdot 6000^3$) and it is almost impossible for the network to memorize the training set. The inferences shown in figure \ref{visual_inference} are additionally performed on images from the test set, and therefore demonstrate true generalization of the task. 

For spike encoding of the images, each pixel will be assigned to a neuron and its firing rate is related to the assigned pixel value $p_i\in [0,1]$ via a factor $r_\mathrm{max}$ by $r_i = p_i r_\mathrm{max}$. For images of size $28 \times 28$, this gives an IO-population size of $784$.

In figure \ref{visual_inference}, can be seen that the network has learned to replace all missing parts of the relation by an output which resembles an artificial stimulus. The network has therefore learned to set the abstract meaning of the images in relationship with each other.

\section{Discussion}

\subsection{Comparison the previous implementations of the relational network}

The network of \cite{Diehl:2016} uses bio-inspired learning algorithms (different variants of STDP), which are applied on populations of inhibitory and excitatory leaky integrate-and-fire neurons. Learning in their implementation is not split into several phases where the roles of the IO-population change, but all populations are treated equivalently during example presentation. This is possible because learning is solely based on correlated activity patterns of neuron. This leads however to a problem during inference. Activity in the network tends to attenuate strongly, since the network was not trained on patterns where no input is provided to one of the populations. While this is not a problem for the rather simple inference on relations of numbers represented by the weighted mean of the output pattern, it might become a problem if the scale of the inferred pattern is relevant for inference, or if the network is very deep and activity dies out completely before it arrives at the inferring population. This problem is partially avoided by using a high number of neurons with self-regulating recurrent connectivity. The high level of recurrent connectivity can however lead to attractor states which are detrimental for learning. This problem can be solved with a wake-sleep type algorithm \cite{Thiele:2017WSA}.  However, the architecture still requires a high number of neurons and careful parameter tuning for good performance. Additionally, although the approach is bio-inspired, it is not necessarily easy to implement in neuromorphic hardware due to the complex nature of the STDP rules and some tricks which are used to stabilize learning (i.e.\ regular weight normalization). 

Our algorithm is much simpler than the more biologically inspired approaches in the sense that it requires less parameters tuning. As can be seen in table \ref{parameters}, the only network parameters which have to be tuned additionally compared to a standard ANN are the threshold values. In contrast to the STDP based approaches, our algorithm optimizes the network using gradient descent on an exact objective function \eqref{loss_function}. One limitation of our approach is that for training and inference the network has to decide in advance which variables it wants to infer and disable the corresponding synapses (as visualized in figure \ref{network_learning}). It therefore requires a kind of attention mechanism. It the simplest case, this attention mechanism can for example be implemented by observing which subset of the populations receives the largest number of input spikes, and disabling all connections which would infer with the feedforward structure corresponding to this inference direction.

\subsection{Biological plausibility}

As all implementations of backpropagation, our learning rule is non-local, in the sense that learning requires the presence of a feedback signal external to the neuron. However, this non-locality exists in any other architecture where the neural ensembles have to process external information which does not directly arrive at the neuron. This includes the STDP-based architecture of \cite{Diehl:2016}, where this information is implicitly communicated by the inter-population connections. Also in our approach, errors are communicated as spikes between populations, which allows us to see this external information simply as another form of special synaptic input that arrives at the neuron at a different time. The advantage of our approach is that spikes have a clear interpretation: they encode an approximation of the backpropagated error. This allows us to use the power of backpropagation while maintaining spike-based communication between populations. The main difference to biological inspired architectures such as \cite{Diehl:2016} is that we require bidirectional synapses if we want to represent the gradient accurately. However, as for other ANN implementations using the backpropagation algorithm, this condition might be lifted by using an approximation based on randomized weights, such as (direct) feedback alignment \cite{Lillicrap:2016}.

Note that the two integrators $V$ and $U$ used in the description of the backpropagation algorithm could even be presented by the same membrane potential, since they are not used at the same time and (in our implementation) they have the same threshold value.

\section{Conclusion and Outlook}

We demonstrated that a neuromorphic version of the backpropagation algorithm can be used to train a network for relational inference. We showed that our implementation can be advantageous compared to previous STDP-based implementations in several aspects. Additionally, we showed that our network is able to learn a visual XOR task based on images of handwritten digits. This could make our approach promising for low power mobile platforms, where several sensor outputs have to be processed and set into relationship which each other. In the work presented here, we focused on relations of stimuli of the same type, but in principle our network could be extended to merge stimuli of different nature, such as visual, audio, or numeric stimuli, as long as they can be represented by firing rates of spiking neurons.

In future work we want to investigate if our approach can be scaled to more complex relationships between stimuli. Additionally, it would be interesting to adapt the algorithm to analog or mixed signal neuromorphic implementations (such as \cite{Moradi:2018}), which would allow processing even closer to the sensor.

\section*{Acknowledgment}

We would like to thank the CapoCaccia Workshop and the researchers at INI, with special attention to Matthew Cook, for the helpful discussions on the principles of the Network of Relations. This work was partly funded by EU grant NeuRAM3 with grant number 687299 and by the ERC consolidator Grant to G.I. NeuroAgents with grant number 724295.



\bibliographystyle{IEEEtran}
\bibliography{IEEEabrv,references}
%

\end{document}